\documentclass{article}

\usepackage[preprint]{neurips_2025}

\usepackage[utf8]{inputenc} % allow utf-8 input
\usepackage[T1]{fontenc}    % use 8-bit T1 fonts
\usepackage{hyperref}       % hyperlinks
\usepackage{url}            % simple URL typesetting
\usepackage{booktabs}       % professional-quality tables
\usepackage{amsfonts}       % blackboard math symbols
\usepackage{nicefrac}       % compact symbols for 1/2, etc.
\usepackage{microtype}      % microtypography
\usepackage{xcolor}         % colors
\usepackage{epsfig}
\usepackage{graphicx}

\title{multi-sensor zero-shot}

\title{Exploring the Versatility of Generalist Multimodal Models for Multi-Sensor Remote Sensing Applications}

\title{Multi-sensor Zero-shot Learning for Generalist Multimodal Models}

\title{Zero-Shot Multi-Spectral Remote Sensing: Repurposing Generalist Multimodal Model Gemini 2.5 for Multi-Spectral Remote Sensing Applications}
\title{Repurposing the Generalist Multimodal Gemini 2.5 Model for Zero-Shot Multi-Spectral Remote Sensing}

\title{Zero-Shot Multi-Spectral Learning: \\ Reimagining a Generalist Multimodal Gemini 2.5 Model for Remote Sensing Applications}

\author{%
 Ganesh Mallya$^\mathsection$ \hspace{1cm} Yotam Gigi$^\dagger$ \hspace{1cm} Dahun Kim$^\mathsection$ \hspace{1cm} Maxim Neumann$^\mathsection$   \\ 
 \And
 Genady Beryozkin$^\dagger$  \hspace{1cm} Tomer Shekel$^\dagger$ \hspace{1cm} Anelia Angelova$^\mathsection$ \\ \\ %\thanks{Use footnote for providing further information about author (webpage, alternative address)---\emph{not} for acknowledging funding agencies.} 
  $^\mathsection$Google DeepMind  \hspace{1.0cm}  $^\dagger$Google Research\\
}

\begin{document}

\maketitle

\begin{abstract}

Multi-spectral imagery plays a crucial role in diverse Remote Sensing applications including land-use classification, environmental monitoring and urban planning. These images are widely adopted because their additional spectral bands correlate strongly with physical materials on the ground, such as ice, water, and vegetation. This allows for more accurate identification, and their public availability from missions, such as Sentinel-2 and Landsat, only adds to their value. Currently, the automatic analysis of such data is predominantly managed through machine learning models specifically trained for multi-spectral input, which are costly to train and support.
Furthermore, although providing a lot of utility for Remote Sensing, such additional inputs cannot be used with powerful generalist large multimodal models, which are capable of solving many visual problems, but are not able to understand specialized multi-spectral signals. 

To address this, we propose a training-free approach which introduces new multi-spectral data in a Zero-Shot-only mode, as inputs to generalist multimodal models, trained on RGB-only inputs. Our approach leverages the multimodal models' understanding of the visual space, and proposes to adapt to inputs to that space, and to inject domain-specific information as instructions into the model. 
We exemplify this idea with the Gemini2.5 model and observe strong Zero-Shot performance gains of the approach on popular Remote Sensing benchmarks for land cover and land use classification and demonstrate the easy adaptability of Gemini2.5 to new inputs.
These results highlight the potential for geospatial professionals, working with non-standard specialized inputs, to easily leverage powerful multimodal models, such as Gemini 2.5, to accelerate their work, benefiting from their rich reasoning and contextual capabilities, grounded in the specialized sensor data.

\end{abstract}

\date{May 2025}

\maketitle

\section{Introduction}

\begin{figure}[t]
   \centering
     \includegraphics[width=\linewidth]{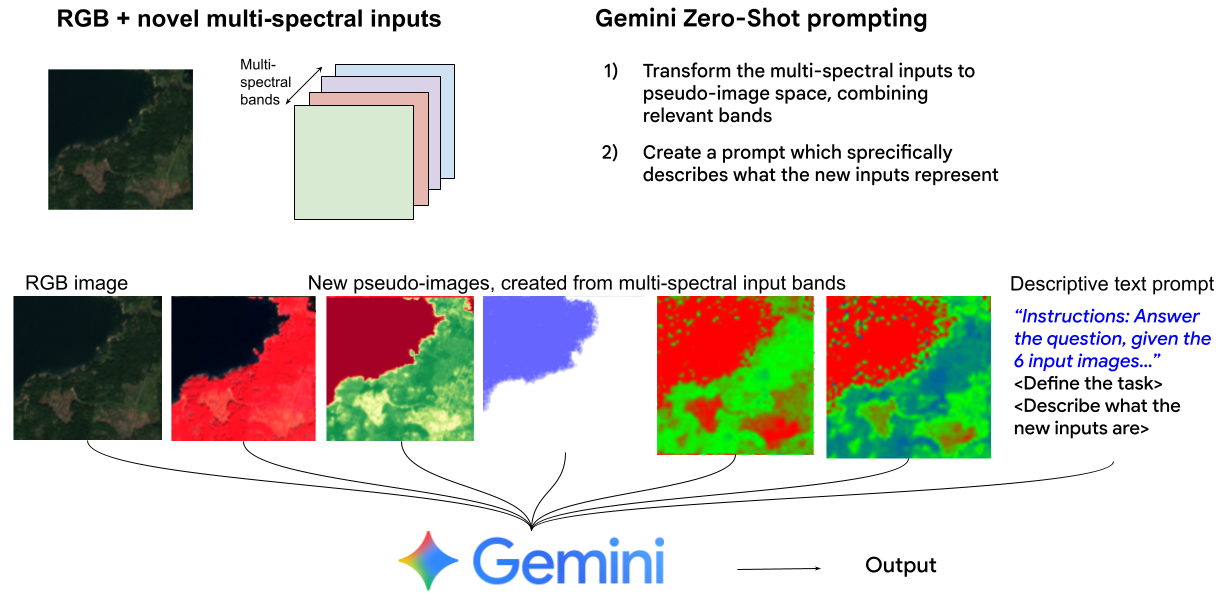}
  %  \includegraphics[width=\linewidth]{figures/Gemini multi-spectral.png}
%   \includegraphics[width=\linewidth]{figures/Teaser.png}
  % \vspace{-6mm}
   \caption{We demonstrate here how a generalist Gemini2.5 multimodal model, trained on RGB-only inputs, can be adapted, when queried Zero-Shot and without any training, to understand new and unfamiliar multi-spectral inputs. This improves Gemini's already strong performance on these tasks and extends its applicability to more Remote Sensing tasks which often rely on extra visual inputs.
   }
   \label{fig:teaser}
 %  \vspace{-5mm}
\end{figure}

%Remote sensing useful applications: land-cover classification, urban planning and development, sustainability, analyzing agricultural landscape and its potential, marine monitoring, forest composition analysis e.g. for carbon footprint and etc

Large Multimodal Models (LMM) are trained on a variety of tasks and serve as generalist learners. They are powerful  visual understanding tools with generalization going beyond the training data, and often exhibit surprising capabilities across many unseen tasks.  
Remote Sensing applications can take advantage of the broad visual understanding capabilities of such large generalist task-agnostic models. However, most generalist multimodal models are available for only RGB images inputs and are not suitable for other input formats or modalities,  e.g. multi-spectral or other inputs, which are common in Remote Sensing and are often available e.g. from  Sentinel-2 and Landsat sensors. 

At the same time, multi-sensor inputs for Remote Sensing are very valuable as they contribute complementary  information, e.g. different input types provide different perspectives on the same data, for example the multi-spectral bands capture additional frequencies which can detect various aspects of the material properties on the ground. Combining multiple sensors can lead to more accurate and detailed responses, whereas individual sensors, e.g. RGB, have blind spots and limitations.  
A common approach is to train domain-specific foundational models~\cite{satmae2022,Prithvi-100M-preprint,SpectralGPT,HyperSigma} which are designed to input these specialized sensing modalities e.g. RGB, NIR and SWIR channels. These models tend to replicate the large multimodal models, but instead train on a collection of sensor-specific data, which is a very costly process and requires domain-specific data collection which is non-trivial for Remote Sensing.

Furthermore, the new input sensors and their input specifications might change in time, which makes it hard to support such specialized foundational models. For example, the data collection is very sensor-dependent which might not be compatible with previous or subsequent sensors: new generation of sensors might extend the number of inputs or produce different or higher quality data than previous ones; new types of sensors can also become available. It will be desirable that the multimodal models do not need to be re-trained for each new change in the input sources. 
The models should be resilient to newer or different data collection efforts and be easily adaptable to sensors with a similar but not exactly the same set of inputs. 
%Handling multiple remote sensing modality inputs is overwhelming, particularly if new re-training needs to happen and also because they might be sensor dependent which might not be compatible with previous ones.

%For example the Remote Sensing domain has multi-sensor  inputs which are very specific to its applications and the LMMs might not have related images in the training data. 

But how can we leverage these generalist multimodal models for specialized tasks without additional training or fine-tuning?
We here propose to re-purpose generalist multimodal models to understand new and unknown input sensing modalities leveraging the generalist model's ability to understand visual data, and the model's ability to understand detailed and specific instructions about the input. This is done without any need for training or model adaptations.
The key idea is to represent the new inputs as pseudo-image inputs and for each of them to contextualize the inputs with extensive text descriptions which specifically describe what the visual embodiment of the inputs represents (Figure~\ref{fig:teaser}).
For example, for a mutli-spectral input of 12 possible bands, we create a number of pseudo-color images, e.g. some of which typically represent NIR, SWIR, NDMI, as well as others which are new. These pseudo images will still be easily understood by the model's visual space. At the same time, with an interpretative detailed text prompt we describe how each of these images is created which bands were included and what each band potentially measures in physical space, as well as, interpret its meaning e.g. is `indicative of the presence of moisture', or `represents green vegetation'.
Our approach takes full advantage of the inherent and powerful visual representations of generalist LMMs but is interpreting them in a way that is helpful for the specific application. 
This makes generalist LMMs much more versatile and useful for Remote Sensing applications.

We demonstrate this idea with the Gemini2.5 model~\cite{Gemini2.5}. %,  in Zero-Shot setting, where the model is not trained on such data. 
We observe strong performance of the model with multi-spectral inputs in an entirely Zero-Shot setting, with large improvements in Zero-Shot performance, as well as, achieving new state-of-the-art (SOTA) performances.
% TODO Add this to future work
These results show that large generalist multimodal models can be easily enabled to understand data from new input sensors, thus improving their performance levels without the need for additional training and further increasing their applicability to many more tasks in Remote Sensing. %This can also be applied to various types of sensor measurements which have corresponding spatial mapping to imagery, e.g. from thermal sensing and others.

\section{Previous work}

Multimodal foundational models are widely used for Remote Sensing~\cite{FoundSurvey}:
These models are specifically trained with Remote Sensing data in order to target these applications. For example,
RemoteCLIP~\cite{RemoteCLIP},
SkyCLIP~\cite{skyscript},
RS-CLIP~\cite{RSCLIP} are commonly used.
Another common approach is to create domain-specific foundational models which are fine-tuned from already pre-existing foundational multimodal models, with data specific to Remote Sensing or to pre-train foundational models for Remote Sensing, e.g.~\cite{SatlasPretrain,SkySense,EarthPT,RemoteCLIP,skyscript,RSCLIP,FoundSurvey,kuckreja2023geochat,Scale-MAE, GFM, Billion}.
Some other approaches proposed variations of light-weight fine-tuning or adaptation, e.g.~\cite{LoRA-NIR}.
The above-mentioned Remote Sensing models, though, all work in the RGB-only domain.
%Some of these models leverage additional remote-sensing pre-training datasets, i.e. pre-trained on Geo-specific data: SkyCLIP~\cite{skyscript}  

Building foundational models for multi-spectral or multi-sensor Remote Sensing data~\cite{SkySense,satmae2022,Prithvi-100M-preprint,SpectralSpatial,hong2022spectralformer,MaskedHyperspectral,SpectralGPT,CROMA,OmniSat,MMEarth} has been also very beneficial, although it is more challenging due to the diverse set of inputs.
SatMAE~\cite{satmae2022} proposes a specialized extension for pre-trained Remote Sensing models where pre-training with multi-spectral imagery is specifically done.
The Prithvi model series~\cite{Prithvi-100M-preprint} have developed foundational models which input the 
Blue, Green, Red, Narrow NIR. SWIR 1, SWIR 2 as input channels.
SkySense~\cite{SkySense} proposes a model for RGB, multi-spectral, as well as, temporal inputs.
In~\cite{Interband} the authors propose training on multi-spectral data by separating the bands according to their resolutions. They present results with M1 60x60m (multi-spectral bands 1, 9), M2 (bands 5, 6, 7, 8A, 11, 12) 20x20m and M3 (bands 2, 3, 4, 8) which are RGB and Near infra-red (NIR) at 10x10 m resolution.
Joint-training across multi-spectral (Sentinel-2 and Sentinel-1) inputs shows more robust performances ~\cite{linial2025Enhancing}. 
Some other foundational models focus on incorporating the rich sensing capabilities of hyper-spectral imagery. For example,
SpectralGPT~\cite{SpectralGPT} proposed a spectral foundational model with 3D Generalized Transformer.
HyperSIGMA~\cite{HyperSigma} propose a sparse attention learning to optimize the spectral-spatial feature extraction across hyper-spectral inputs.
While more comprehensive in their input modalities, these foundational models are not flexible in terms of adding new input channels. Furthermore training multi- or hyper-spectral models is a very resource-intensive task.

In contrast, our work leverages generalist (RGB-based) large multimodal models which are not necessarily specific to Remote Sensing, neither in pre-training data nor in input modalities. We show how that such multimodal models, e.g. here Gemini 2.5, can accommodate various multi-spectral new and unfamiliar inputs in a Zero-Shot fashion, i.e. without any model training or adaptations needed.

%and GeCo are foundational models for agriculture.
%RSP~\cite{} is a foundational model with multi-spectral data for agriculture (precision agriculture), crop monitoring.

% has some more models
%https://arxiv.org/pdf/2507.05390

%Retrieval Augmentation:
%~\cite{RANGE}

%Prompting RSPrompter learning prompting for segmentation in RS applications~\cite{RSPrompter}

%~\cite{Generalizability}

Another type of approach is to learn semantically meaningful embeddings from large number of Remote Sensing inputs.
Remote Sensing embeddings treat prediction tasks as a two-step process, a featurization step, which transforms input multi-sensor inputs into learned compact features, or emebeddings, and a regression step which associates the embeddings to the outcomes or labels per tasks, for example MOSAIKS~\cite{MOSAIKS}, SatlasPretrain~\cite{SatlasPretrain},
EarthPT~\cite{EarthPT}, %is an Earth observation foundational model
S2Vec~\cite{s2vec},
AlphaEarth Embeddings~\cite{AlphaEarth}. These features can integrate various input sensors, but are appropriate for clustering, further  fine-tuning, of low-shot learning scenarios. They are not easy to use in zero-shot settings, as our model is.

\section{Multi-Sensor Zero-Shot Inference}

Generalist multimodal models are trained on images harvested from the web. As a result, their performance on natural images in Zero-Shot settings is quite impressive. However, they are not able to understand additional inputs, such as, for example, multi-spectral inputs, commonly used for Remote Sensing.

In this section we demonstrate how we can adapt these models to other sensing modalities, still performing Zero-Shot inference, and without the need to re-train these models.

We propose a simple adaptation of the new and unfamiliar inputs and a specialized informative prompt explaining the process of input generation, by which we easily turn a generalist multimodal model into a model which understands and performs well on new multi-sensor data (Figure~\ref{fig:teaser}).
The key idea is to combine the multi-spectral inputs in various ways and represent them as images or pseudo-image inputs, and for each of them, to contextualize the inputs with extensive prompts, i.e. text descriptions which specifically describe what the inputs visual embodiment represents. 

\begin{figure}
    \centering
    \includegraphics[width=0.15\textwidth,angle=90]{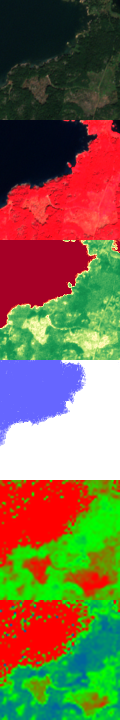}
    \includegraphics[width=0.15\textwidth,angle=90]{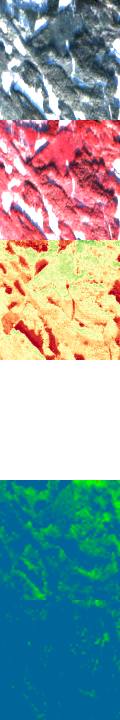}
    \includegraphics[width=0.15\textwidth,angle=90]{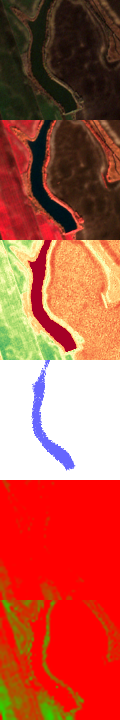}
    \includegraphics[width=0.15\textwidth,angle=90]{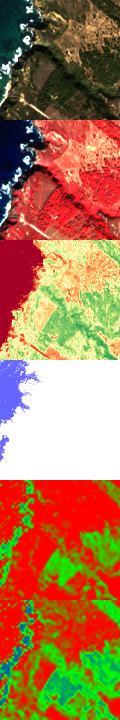}
    \includegraphics[width=0.15\textwidth,angle=90]{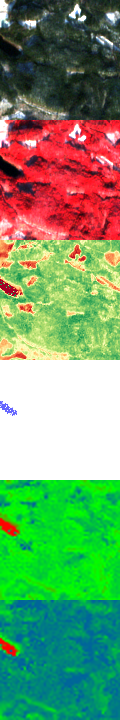}
    \includegraphics[width=0.15\textwidth,angle=90]{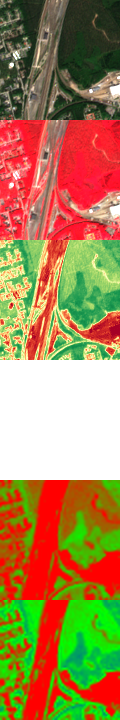}
    \includegraphics[width=0.15\textwidth,angle=90]{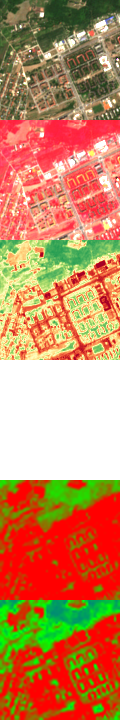}
    \caption{Examples of the six inputs from BigEarthNet, which is a multi-spectral dataset. The inputs used here are RGB images (Column 1), and five different multi-spectral inputs, represented here as pseudo-color images, so that the model can understand them. Column 2 is constructed from the 8th multi-spectral band, which contains a broader range of the visible and near-infrared spectrum and the 4th and 3rd bands. Column 3 contains an a pseudo image, similar to NDVI, which is sensitive to vegetation. Column 4 shows a set of multi-spectral inputs which are more sensitive to water and Columns 5 and 6 might be also helpful for vegetation.
     With that simple adaptation and a specialized prompt explaining the process of input generation we easily turn a generic foundational model into a model understanding and performing well on multi-sensor Remote Sensing data.
     %Images shown are from the BigEarthNet dataset.
    }
    \label{fig:ben_examples}
\end{figure}

\begin{figure}
    \centering
    \includegraphics[width=0.15\textwidth,angle=90]{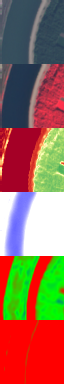}
    \includegraphics[width=0.15\textwidth,angle=90]{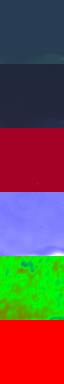}
    \includegraphics[width=0.15\textwidth,angle=90]{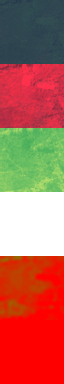}
    \includegraphics[width=0.15\textwidth,angle=90]{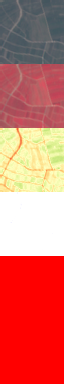}
    \includegraphics[width=0.15\textwidth,angle=90]{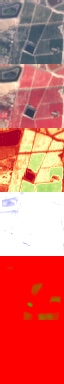}
    \includegraphics[width=0.15\textwidth,angle=90]{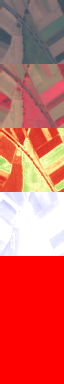}
    \includegraphics[width=0.15\textwidth,angle=90]{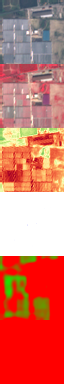}
    \caption{Examples of the six inputs which are obtained from the multi-spectral EuroSat dataset and represented here as pseudo-color images. See Figure~\ref{fig:ben_examples} for description of the images.}
    \label{fig:eurosat_examples}
\end{figure}

For example, for a multi-spectral input of 12 possible bands, we create a number of pseudo-color images which typically represent NIR, SWIR, NDMI, and others, which are common and can be helpful for Remote Sensing tasks. 
Example pseudo-image inputs, generated from various multi-spectral bands, are visualized in Figure~\ref{fig:ben_examples} for the BigEarthNet~\cite{BigEarthNet} dataset, and in Figure~\ref{fig:eurosat_examples} for the EuroSat~\cite{helber2019eurosat} dataset.

The above-mentioned newly-generated images are provided as inputs to the model. Then, with an descriptive and detailed prompt we describe how each of these images is created, that is, which bands were included and what each band potentially measures in physical space, as well as, interpret its meaning via a text prompt. % e.g. `this band is indicative of the presence of moisture', or `represents green vegetation'. 
In this way the multi-spectral input modalities can still be represented as image inputs which the generalist multimodal models are able to understand, providing a format which is well suited for image representations. Furthermore, the text prompt provides the necessary information and context about the new inputs.
For example, we can list the input sensors within the prompt: `The images are generated by selecting different bands of the Sentinel-2 satellite. 
The band information is as follows: 
\begin{itemize}
    \item 1. B02: Blue band at 10-meter resolution
    \item 2. B03: Green band at 10-meter resolution
    \item 3. B04: Red band at 10-meter resolution
\item 4. B05: Red edge band (Central wavelength of 704.1 nm) at 20-meter resolution
\item 5. B06: Red edge band (Central wavelength of 740.5 nm) at 20-meter resolution
\item 6. B07: Red edge band (Central wavelength of 782.8 nm) at 20-meter resolution
\item 7. B08: NIR band at 10-meter resolution
\item 8. B8A: Narrow NIR band at 20-meter resolution
\item 9. B01: Costal Aerosol band at 60-meter resolution
\item 10. B09: Water vapor band at 60-meter resolution
\item 11. B11: SWIR band (Central wavelength 1613.7 nm) at 20-meter resolution
\item 12. B12: SWIR band (Central wavelength 2202.4 nm) at 20-meter resolution.'
\end{itemize}

Using this information,
we can further describe the inputs within the prompt, as follows (please see the Appendix for the full prompt):  
%\begin{figure}[t]
%   \centering

%   \includegraphics[width=\linewidth]{figures/6inputs.png}
 %    \includegraphics[width=\linewidth]{figures/BigEarthNet-example-water.png}
  % \vspace{-6mm}
  % \caption{Examples of the six inputs which are obtained from teh multi-spectra data and represented here as pseudo-color images so that the model can understand them. With that simple adaptation and a specialized prompt explaining the process of input generation we easily turn a generic foundational model into a model performing very well on multi-sensor remote sensing data. %TODO: Will swap this out with a few more informative examples which also show water bodies so that the 4th image is meaningful. 
   %Bottom: to redo shows examples of NDVI (vegetation index) and NDWI (water index) images, todo will swap them with propoer full examples. 
   %}
%   \label{fig:examples}
 %  \vspace{-5mm}
%\end{figure}

%Additional inputs: 

\begin{itemize}
    \item  The first image is the RGB image generated using B04, B03 and B02 bands. 
    \item The second image is the False Color Composite image generated using B08, B04 and B03 bands. 
    \item The third image is the NDVI image which is a colormap with values Red, Yellow and Green generated using B08 and B04 bands.  \item The fourth image is NDWI image whose values are in the range of -0.8 to 0.8 and using a color map varying linearly as [(1, 1, 1), (1, 1, 1), (0, 0, 1)]. 
    \item The fifth image is the NDMI images generated using B8A and B11 bands with the colormap varying linearly as [(1, 0, 0), (0, 1, 0), (0, 0, 1)].
    \item The sixth image is the NDMI images generated using B8A and B12 bands with the colormap varying linearly as [(1, 0, 0), (0, 1, 0), (0, 0, 1)].
\end{itemize}

%In this way multi-sensor modalities can still be represented as image inputs which generalist models are able to understand, providing a format which is suited for image representations.
This approach takes full advantage of the inherent powerful visual representations of generalist LMMs but is interpreting the new visual input in a way that is helpful for the task at hand and is specialized to the inputs of the Remote Sensing applications. 

\subsection{Implementation details}
For concreteness, we here create five additional input pseudo images, in addition to the standard RGB input. They are as follows: 1) a false color composite image generated using bands B08, B04 and B03 bands; 2) an NDVI image using bands B08 and B04 bands; 3) an NDWI image 4) an NDMI image using bands B8A and B11; 5) an NDMI images generated using B8A and B12 bands. Color maps are applied when necessary.
When generating the true color image and false color composite image (i.e. the first and the second images), we perform normalization of each band to bring the input value in the range [0, 1]. We then scale these values to fall in the range [0, 255] before stacking the bands together to be saved as RGB image.

%Many more combinations are possible as shown here: https://gisgeography.com/sentinel-2-bands-combinations/

\section{Experiments}

%\subsection{Multi-spectral Zero-Shot Results on BigEarthNet}
\subsection{Zero-Shot Results with Multi-Spectral Inputs on BigEarthNet}

We first show the performance of the proposed approach on the popular BigEarthNet dataset~\cite{BigEarthNet}, which is a land cover classification dataset which provides 12 %(or 13) 
multi-spectral bands.
The dataset is a multi-label dataset, which means that more than one label are considered correct per example. For that reason the main metrics in evaluating this dataset is the F1 metrics, which is also consistent with performance reporting in the literature~\cite{Zhang2024GoodAt}. We also report Precision and Recall values to gain understanding of these components and their contribution to the final F1 metrics.
Furthermore, since the dataset is multi-label, we prompt the model specifying that more than one class is possible as an output. We do not limit the number of output classes for the model. See the Appendix for the specific prompt.

We here report Zero-Shot results, demonstrating the performance of Gemini 2.5, without the need for any fine-tuning for the new multi-spectral inputs or to a specific task or dataset. Our results are aligned with recent work~\cite{Zhang2024GoodAt} which benchmarked GPT-4V and several other methods in the Zero-Shot setting, as well.

\begin{table}[ht]
    \centering
    \begin{tabular}{l|c|cc}
    \toprule
    Model & F1 & Precision &Recall \\
    \midrule
   Gemini2.5 with RGB  &0.388  &0.516 &0.311 \\
   Gemini2.5 with Multi-Spectral (Ours) &\textbf{0.429}  &0.526 &0.363\\
   %Gemini2.5 with RGB  &0.406 &\textbf{0.560} &0.318 \\
   %Gemini2.5 with Multi-spectral (Ours) &\textbf{0.465} &0.554 &\textbf{ 0.400} \\
    \midrule
    \bottomrule
    \end{tabular}
    \caption{BigEarthNet land cover classification Zero-Shot results (multi-label, multi-class). The same base multimodal model is used without modifications (Gemini 2.5). The main performance metrics is F1. The gain realized from multi-spectral inputs is significant by \textbf{+0.04} in F1. Classification with 43 classes.}
    \label{tab:zs_comp}
\end{table}

\begin{table}[h]
    \centering
    \begin{tabular}{l|c|cc}
    \toprule
    Model & F1 & Precision &Recall \\
    \midrule
 % 43 classes old results, 43 classes moved to another table
 %  Gemini2.5 with RGB  &0.406 &\textbf{0.560} &0.318 \\
  % Gemini2.5 with Multi-spectral (Ours) &\textbf{0.465} &0.554 &\textbf{ 0.400} \\
     \midrule
     % 19 classes old results
  % Gemini2.5 with RGB  &0.410 &0.533 &0.334 \\
  % Gemini2.5 with Multi-spectral (Ours) &\textbf{0.476} &\textbf{0.568} &\textbf{ 0.409} \\
   Gemini2.5 with RGB  &0.414  &0.542 &0.335 \\
   Gemini2.5 with Multi-Spectral (Ours) &\textbf{0.453}  &0.566 & 0.377 \\
   % \textbf{+0.039}
   \midrule
   GPT-4V~\cite{Zhang2024GoodAt}   &0.38  &0.49 &0.43 \\
Qwen-VL-Chat~\cite{Zhang2024GoodAt} &0.40  &0.57 &0.39  \\
InstructBLIP-FLAN-T5-xxl~\cite{Zhang2024GoodAt} &0.02  &0.41 &0.01  \\
InstructBLIP-Vicuna-13b~\cite{Zhang2024GoodAt}  &0.01  &0.01 &0.06\\
LLaVA-v1.5~\cite{Zhang2024GoodAt}  &0.39  &0.27 &0.83 \\
    \midrule
    \bottomrule
    \end{tabular}
    \caption{BigEarthNet land cover classification Zero-Shot results (multi-label, multi-class). The same base multimodal model is used without modifications (Gemini 2.5). The main performance metrics is F1. The gain realized from multi-spectral inputs is about \textbf{+0.04} in F1. Classification with with 19 classes. At the bottom, we compare to SOTA Zero-Shot results on the 19-classes variant and observe a \textbf{+0.053} gain in F1 over the best SOTA results.}
    \label{tab:zs_comp19}
\end{table}

Table~\ref{tab:zs_comp} shows the gains obtained from the proposed multi-spectral approach, compared to RGB-only. In both cases we report Zero-Shot performances and use the same Gemini 2.5 model for inference, where the same model is used for both RGB and multi-spectral inputs.
 We see that multi-spectral inputs provide significant improvements, and are realized without needing any additional training or adaptation. 
The results reported are for the original version of the dataset with 43 target categories.

Since some of the original classes can be quite ambiguous and have overlapping semantic meaning, a newer version of the dataset is proposed with 19 classes. In Table~\ref{tab:zs_comp19} we show the results on BigEarthNet with 19 classes, where the 43 classes are merged into fewer and less ambiguous classes, as adopted in the literature~\cite{BigEarthNet19Cl,Zhang2024GoodAt}.
As seen, the proposed approach performs consistently well, with similar margins of improvements, on this version of the dataset, as well.

Table~\ref{tab:zs_comp19} (bottom) shows the Zero-Shot results compared to the state-of-the-art (SOTA) Zero-Shot  methods. As seen, Gemini 2.5 already has very strong performance in the RGB-only setting, compared to powerful other models. Its performance is further improved by the proposed multi-spectral approach.
%
%We compare results of Zero-Shot performance with multi-spectral and RGB-only. %For context, we provide the upper-bounds received with full fine-tuning of this dataset in prior works.

\subsection{Zero-Shot Results with Multi-Spectral Inputs EuroSat}

We further evaluate the performance of our Zero-Shot inference-only approach on the popular EuroSat land use classification benchmark~\cite{helber2019eurosat}. This benchmark is of larger resolutions RGB imagery. It has 10 classes and thus accuracy is its main performance metric.
Table~\ref{tab:zs_eurosat} shows the results. 
As seen, with our proposed approach, we can obtain +3\% accuracy gains on this benchmark. 

We report Zero-Shot results for this benchmark, as well as, SOTA Zero-Shot results, including the results of ZLaP which uses more information. More specifically ZLaP uses {\bf Inductive Zero-Shot inference}, meaning it is allowed to use the dataset examples, but without labels. As seen, our approach, outperforms the SOTA approaches, and is preferred to this method, as well.

While this dataset obtains high performance when fine-tuned, with classification accuracies in the 90s, we see here that zero-shot performance has a larger gap. The reason might stem from the fact that some classes are hard to distinguish without any a priori information e.g. `Annual Crop' vs `Permanent Crop' and the model cannot distinguish between them without an association of how each one might look visually. Fine-tuning is helpful here as it allows a machine learning model to associate a visual input to a specific class label, which is not available in the Zero-Shot setting.   We also note that there are also some noisy and mislabeled examples in this dataset. Nevertheless, we observe improved performance with good accuracy gains of our approach for this benchmark, as well, and are able to decrease the gap to the fully-fine-tuned performance further.

\begin{table}[]
    \centering
    \begin{tabular}{l|c}
    \toprule
    Model & Classification Accuracy, Top 1 (\%) \\
     \midrule
%    \midrule
%  Gemini2.5 with RGB &65.20 \\    
%  Gemini2.5 with Multi-Spectral (Ours) &\textbf{68.21} \\
Gemini2.5 with RGB & 66.3 \\    
 Gemini2.5 with Multi-Spectral (Ours) &\textbf{69.1} \\
    \midrule
    CLIP~\cite{radford2021clip} (CLIP-VIT L/14-336px) &59.6 \\
    CLIP~\cite{radford2021clip} (CLIP-VIT L/14) &59.9 \\
    ZLaP~\cite{ZLaP} (Inductive inference) &63.2 \\
    \midrule
    \bottomrule
    \end{tabular}
    \caption{Zero-Shot classification on the EuroSat benchmark. As seen, our proposed approach with inference-only multi-spectral inputs performs best. We further report SOTA result and the best results of ZLaP which uses \textbf{Inductive Zero-Shot inference}, meaning it is allowed to use the dataset examples, but without labels. Our approach, which is Zero-Shot only, outperforms this method too.}
    \label{tab:zs_eurosat}
\end{table}

\subsection{Ablations}

\begin{table}[]
    \centering
    \begin{tabular}{l|c|cc}
    \toprule
    Model & F1 & Precision &Recall \\
     \midrule
RGB only 
&0.406 & 0.560 & 0.318            \\
RGB + NDVI 
&0.415 & 0.509 & 0.350 \\
RGB + NDVI + NDWI
&0.437 & 0.535 & 0.369 \\
RGB + All Multi-Spectral
&0.462 & 0.557 & 0.394    \\
\midrule
RGB only
&0.410 & 0.533 & 0.334     \\
RGB + NDVI
&0.448 & 0.548 & 0.379 \\
RGB + NDVI + NDWI
&0.442 & 0.538 & 0.375 \\
RGB + All Multi-Spectral
&0.479 & 0.565 & 0.415     \\
   \midrule
   \bottomrule
    \end{tabular}
    \caption{Ablation results with adding NDVI and NDWI channels. The experiments are conducted on the BigEarthNet multi-class multi-label benchmark, conducted on a subset of the dataset. Zero-Shot classification. Top: results with 43 classes, Bottom: results with 19 classes.}
    \label{tab:big_earth_ablations}
\end{table}

In Table~\ref{tab:big_earth_ablations} we show an ablation of the Zero-Shot performance of the model, when varying the multi-spectral inputs. The experiment is conducted on a smaller subset of the multi-label multi-class BigEarthNet dataset (on 1000 images only).
As seen, adding all multi-spectral images is performing best. 
Secondly, the performance when including NDVI images alone (together with RGB) is responsible for a good portion of these gains, but is not as good as when using all inputs. 

% These are old results and clearly show the prompt matters but we need to redo them
% The prompt also matters - prompting with the RGB-only prompt is lower (F1 0.427).

%Furthermore, we see that the prompt also matters, where we observe that prompting with straightforward information, even without having to describe additionally what these channels do, seems to work better. We note that even this prompt has sufficient information for the model to use (prompts are shown in the appendix)
%TODO: Add another ablation with RGB-like prompts: Test whether Gemini internally has this information about channels and their frequencies and use-cases or it helps to provide it "in place" by the prompt. 

%\begin{figure}[t]
 %  \centering
  % \includegraphics[width=\linewidth]{figures/BigEarth_Ablation.png}
  % \caption{Ablations on BigEarthNet over the Multi-Spetral inputs as well as the prompts: All image inputs seem to work best. 
  % }
  % \label{fig:big_earth_ablations}
 %  \vspace{-5mm}
%\end{figure}

\subsection{Visualizations}

\begin{figure}[t]
   \centering
     \includegraphics[width=0.15\linewidth]{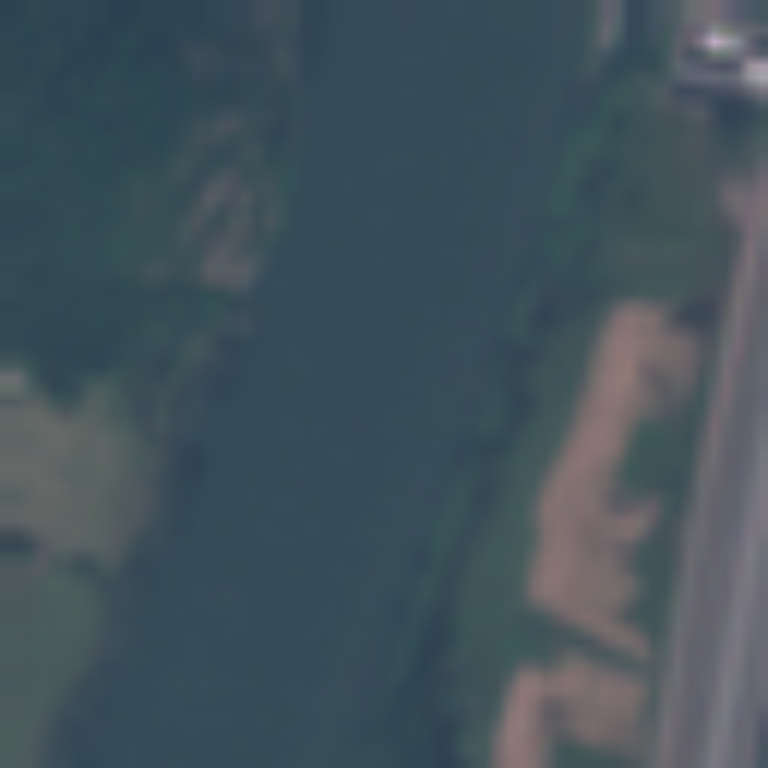}
     \includegraphics[width=0.15\linewidth]{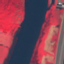}
     \includegraphics[width=0.15\linewidth]{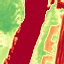}
     \includegraphics[width=0.15\linewidth]{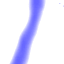}
      \includegraphics[width=0.15\linewidth]{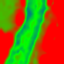}
     \includegraphics[width=0.15\linewidth]{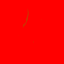} \\
     \includegraphics[width=0.15\linewidth]{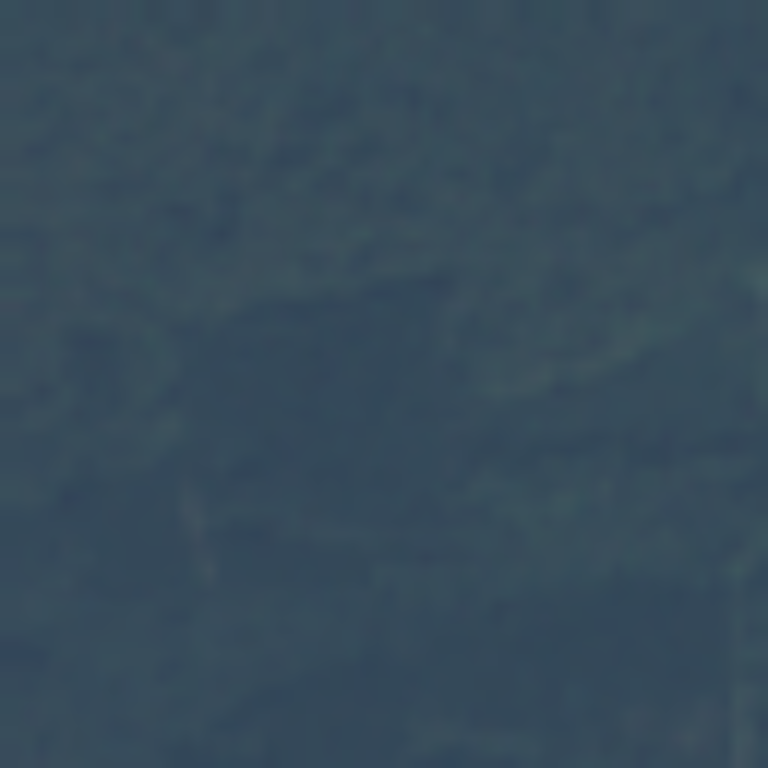}
     \includegraphics[width=0.15\linewidth]{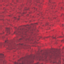}
     \includegraphics[width=0.15\linewidth]{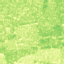}
     \includegraphics[width=0.15\linewidth]{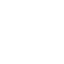}
      \includegraphics[width=0.15\linewidth]{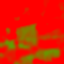}
     \includegraphics[width=0.15\linewidth]{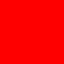} 
     
 %  \includegraphics[width=0.15\linewidth]{figures/ocean_ms_1.png}
 %  \includegraphics[width=0.15\linewidth]{figures/ocean_ms_2.png}
 %  \includegraphics[width=0.15\linewidth]{figures/ocean_ms_3.png}
 %  \includegraphics[width=0.15\linewidth]{figures/ocean_ms_4.png}
 %  \includegraphics[width=0.15\linewidth]{figures/ocean_ms_5.png}
 %  \includegraphics[width=0.15\linewidth]{figures/ocean_ms_6.png}
  % \vspace{-6mm}
   %\caption{Examples of where the multi-spectral inputs are helping for improved performance. The answer provided by the multi-spectral inputs is `Sea and ocean' (37), which is the correct answer. Whereas the answer for the rgb-only input is blank i.e. refuse to answer. As seen by the multi-spectral bands, while the input rgb image is dark with no visible features, the multi-spectral bands are able to detect water (bottom, left-most image which is an NDWI i.e. Normalized Difference Water Index image which is specifically targeting water bodies). We can also note some possible wave-like structures These additional targetted inputs and additional visible visual structures help the model provide the correct answer.
   
    \caption{Example results on the EuroSat data. 
    The multi-spectral inputs are helping for improved performance. 
   \textbf{Top:} The answer provided by the proposed model with multi-spectral inputs is `River' (9), which is the correct answer. Whereas the answer for the RGB-only input is `Forest' (2).
   As seen by the multi-spectral bands, while the input RGB image has blue/green visible features, the multi-spectral bands are able to detect water (bottom, left-most image which is an NDWI i.e. Normalized Difference Water Index image which is specifically targeting water bodies). %We can also note some possible wave-like structures These additional targeted inputs and additional visible visual structures help the model provide the correct answer.
\textbf{Bottom:} An example of `Forest' (2) which the multi-spectral model identifies correctly.
The baseline model is mistaken focusing on the green/blue and classifies it as a 'Sea lake' (10) which is incorrect.
   }
   \label{fig:results_examples}
 %  \vspace{-5mm}
\end{figure}

Figure~\ref{fig:results_examples} shows an example of how the model performs in Zero-Shot with multi-spectral inputs and where differences to RGB-only are seen.
As seen, the additional multi-spectral inputs are crucial to help the model understand situations where the RGB-only image might be ambiguous, e.g. an image with forest which has deep blue/green hue can be often confused with an image with water i.e. sea or ocean regions and vice versa. 
The presence of inputs from multi-spectral bands help with better outcomes.
Specifically, at the top where an image of a river is seen, the RGB-only model predicts a `Forest'.  The answer provided by the proposed model with multi-spectral inputs is `River', which is the correct answer.
At the bottom we see an example of a forest area, which the multi-spectral model identifies correctly.
The baseline RGB-only model classifies it as a 'Sea lake' which is incorrect, likely confusing green/blue colors in the RGB image.

\begin{figure}[t]
   \centering
     \includegraphics[width=0.15\linewidth]{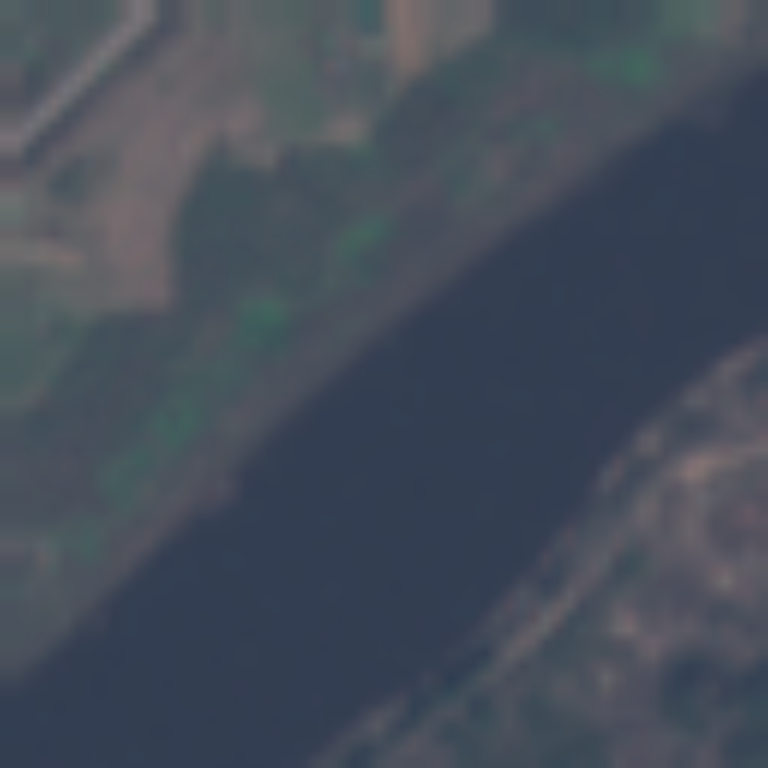}
     \includegraphics[width=0.15\linewidth]{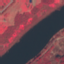}
     \includegraphics[width=0.15\linewidth]{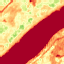}
     \includegraphics[width=0.15\linewidth]{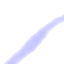}
      \includegraphics[width=0.15\linewidth]{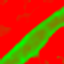}
     \includegraphics[width=0.15\linewidth]{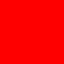}
     
     \includegraphics[width=0.15\linewidth]{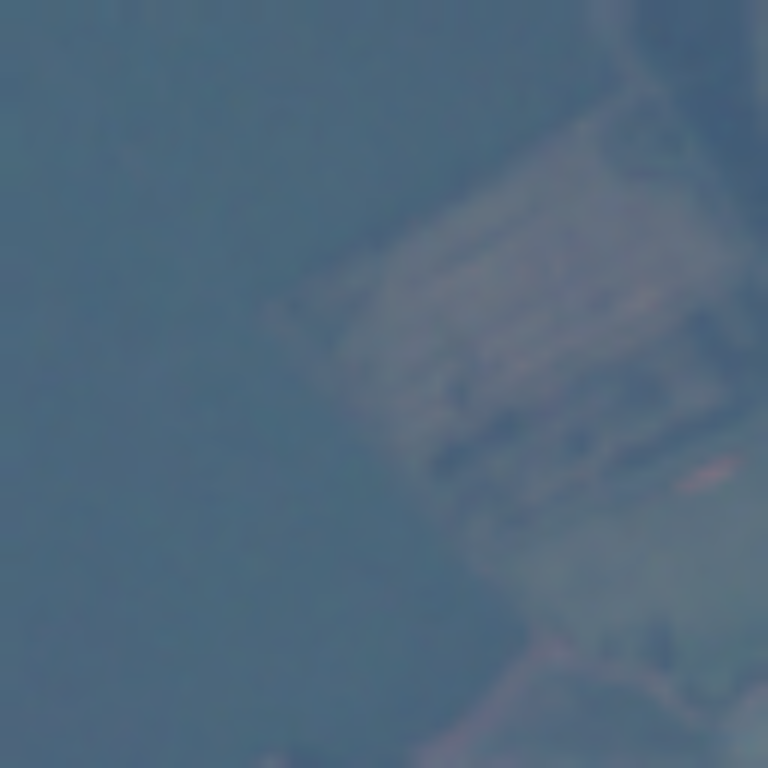}
     \includegraphics[width=0.15\linewidth]{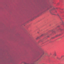}
     \includegraphics[width=0.15\linewidth]{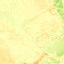}
     \includegraphics[width=0.15\linewidth]{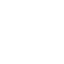}
      \includegraphics[width=0.15\linewidth]{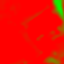}
     \includegraphics[width=0.15\linewidth]{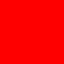}

    \caption{Example results on the EuroSat data where both models are in agreement.
\textbf{Top:} Another example of `River' (9) which both models identify correctly.
\textbf{Bottom:} Both models are mistaken, the proposed model identifies this as 'Annual crop' (1), whereas the baseline model classifies it as a 'Sea lake' . The ground truth is `Pasture' (6).
As compared to Figure~\ref{fig:results_examples}, it is quite challenging to correctly identify the only from RGB image.
   }
   \label{fig:results_examples_same}
 %  \vspace{-5mm}
\end{figure}

Figure~\ref{fig:results_examples_same} further shows examples where both models are in agreement. It shows an example of a river image where both models are correct, here comparing to the top image of Figure~\ref{fig:results_examples}, we see that the RGB input is sufficient, as the colors a much less ambiguous.
An example where both models are mistaken is shown in the bottom image. We note that across the examples in both Figure ~\ref{fig:results_examples} and ~\ref{fig:results_examples_same}, the prediction using the RGB input alone might be quite hard, thus the multi-spectral inputs are key in more accurate understanding. 

%\subsection{Implementation details.} TODO

\subsection{Limitations}
As indicated in this work, we are leveraging the ability of generalist multimodal models to understand visual information, as well as, the associated text prompts, and propose to enrich and adapt the models' inputs in order to expand their multi-modal capabilities to additional sensors. 
Many types of aerial Earth observation data, e.g. overhead Remote Sensing imagery, multi-spectral, hyper-spectral, thermal, LiDAR, Radar, and SAR inputs, all comply with visual input and are particularly fitting for the proposed approach. However, %other types of Earth observation inputs might not be particularly fitting for this approach. For example, 
this might not apply to other sources of scientific or Earth observation data as other types of Earth observations may not necessarily map well to visual inputs. 
%It is an open problem how to include more diverse types of inputs. 
%(TODO: is there such data aerial (e.g. chemical soil composition, which is is an earth observation data but needs labs analysis to obtain, 
%e.g. soil organic carbon content which is measured from the soil itself and obtained via lab analysis.
%; can be mapped to remote sensing for some predictions but training is needed), for the more general case add examples?
% (well these models should read time series, which is easy to add, but the problem is that time series can express various things for which training data or "what means what" is needed).
Furthermore, we note that the model outputs could be sensitive to the text prompt and the order of prompting, and some changes in the prompt might yield different results. %We leave this exploration for future work, as well. 

\section{Conclusions and future work}

Remote Sensing or aerial Earth observation models require lots of resources and datasets for pre-training. Particularly if these models need to be extended to capture multiple additional sensors e.g. multi-spectral or depth information. %, SAR and etc. 
We propose to leverage already pre-trained generic multimodal models, trained on RGB-only, and to adapt them, on-the-fly in a Zero-Shot setting, to the additional new sensing modalities e.g. the multi-spectral imagery, a modality type the models have not seen in training.

To exemplify the idea we use the generalist Multimodal model, here Gemini 2.5, which is trained on vision tasks with RGB-only inputs, and %(typically not related to Remote Sensing) and, importantly, 
which has not used any specialized multi-spectral data during training.
We show that one can very easily leverage the visual understanding of the model and its prompting capability to easily adapt the model to these new unseen inputs and unfamiliar tasks.
We observe improved performance for the Gemini 2.5 model, which already performs very strongly on these RemoteSensing tasks.

We explored several inputs from several multi-spectral bands, with successful results. Other types of sensors  e.g. SAR, depth, or heat sensors, etc ,can also be naturally mapped to visual inputs, and the approach will not need any changes, in order to use them. However, as noted above, there are sensors which may not be particularly fitting. We leave these explorations for future work. 
We further note that out of 12 multi-spectral bands we only created five example image combinations, whereas many other useful combinations can be included. This can improve performance further, without needing to change the approach.

%We demostrate this on three popular multi-spectral benchmarks (BigEarthNet, fMoW, EuroSat) and TODO.

The implications of the proposed work are that such easy-to-use multimodal adaptation of generalist large multimodal models, provides an alternative to the hard-to-generalize and expensive multimodal Remote Sensing training, as well as, for the independent and costly training with specialized Remote Sensing-specific inputs which might be hard to support.
%foundational models with Remote-Sensing domain-specific data which is hard to harvest.
Instead, the proposed approach leverages more general pre-training sources.

%We explored multi-spectral inputs as well as depth sensors, additional types are also feasible e..g hyper-spectral, or depth/SAR, heat sensors etc. They are all particulalrly fitting, and the approach will not need any changes, as all of them map naturally to some visual inputs.
%We note that out of 12 multi-spectral bands we only created 5 example images (combinations), whereas many other useful combinations can be included. This can improve performance further, without needing to change the approach.

%As future work, we can explore expanding to more combinations of inputs, which are not commonly done, e.g. overlays of visible spectral bands e.g. RGB with depth in 1 image or multi-spectral with heat sensing channels. This also is a straightfroward extension.  

%Also as future work, extensions which include changes in time are trivial 

{
\small

\bibliographystyle{unsrt}
\bibliography{main}
}

\appendix

\section{Additional information on example Zero-Shot prompts}

We here provide the prompts used for Multi-Spectral inputs, followed by RGB-only prompt examples. The first prompt examples are for the BigEarthNet, and the last one is for EuroSat. For a different benchmark, the set of available classes will need to change. Since BigEarthNet is a multi-label multi-class dataset, we ask for multiple answers, i.e. by adding `Select all that apply', which is simply  removed for classification datasets, e.g. for EuroSat.

\textbf{Prompt for Multi-spectral inputs on BigEarthNet:}
\begin{verbatim}Instructions: Answer the question asked after the given 6 images of the same scene. 
The images are generated by selecting different bands of the Sentinel-2 satellite. 
The band information is as follows: 
1. B02: Blue band at 10-meter resolution
2. B03: Green band at 10-meter resolution
3. B04: Red band at 10-meter resolution
4. B05: Red edge band (Central wavelength of 704.1 nm) at 20-meter resolution
5. B06: Red edge band (Central wavelength of 740.5 nm) at 20-meter resolution
6. B07: Red edge band (Central wavelength of 782.8 nm) at 20-meter resolution
7. B08: NIR band at 10-meter resolution
8. B8A: Narrow NIR band at 20-meter resolution
9. B01: Costal Aerosol band at 60-meter resolution
10. B09: Water vapor band at 60-meter resolution
11. B11: SWIR band (Central wavelength 1613.7 nm) at 20-meter resolution
12. B12: SWIR band (Central wavelength 2202.4 nm) at 20-meter resolution. 
The first image is the RGB image generated using B04, B03 and B02 bands. 
The second image is the False Color Composite image generated using B08, B04 and 
B03 bands. 
The third image is the NDVI image which is a colormap with values Red, Yellow and Green 
generated using B08 and B04 bands. 
The fourth image is NDWI image whose values are in the range of -0.8 to 0.8 and using 
a color map varying linearly as [(1, 1, 1), (1, 1, 1), (0, 0, 1)]. 
The fifth image is the NDMI images generated using B8A and B11 bands with the colormap 
varying linearly as [(1, 0, 0), (0, 1, 0), (0, 0, 1)]. This image is used for 
identifying wet and dry areas.
The sixth image is the NDMI image generated using B8A and B12 bands with the colormap 
varying linearly as [(1, 0, 0), (0, 1, 0), (0, 0, 1)]. This image is used for 
identifying wet and dry areas.
Output format: Output the option numbers corresponding to the correct answer in the
format "(X)" where X is the correct number choice (among 1 to 43). In case of multiple
correct answers, output the answer choices in the same format separated by commas.
For example, if the correct answers are choices 1 and 3, output "(1),(3)". Only output
the option numbers in the above format and DO NOT OUTPUT the full answer text or any
other extra words.

Question: To which of the following classes do the given images belong to? 
Select all that apply. Possible answer choices:
(1)Agro-forestry areas 
(2)Airports 
(3)Annual crops associated with permanent crops 
(4)Bare rock 
(5)Beaches, dunes, sands 
(6)Broad-leaved forest 
(7)Burnt areas 
(8)Coastal lagoons 
(9)Complex cultivation patterns 
(10)Coniferous forest 
(11)Construction sites 
(12)Continuous urban fabric 
(13)Discontinuous urban fabric 
(14)Dump sites 
(15)Estuaries 
(16)Fruit trees and berry plantations 
(17)Green urban areas 
(18)Industrial or commercial units 
(19)Inland marshes 
(20)Intertidal flats 
(21)Land principally occupied by agriculture, with significant areas of natural vegetation 
(22)Mineral extraction sites 
(23)Mixed forest 
(24)Moors and heathland 
(25)Natural grassland 
(26)Non-irrigated arable land 
(27)Olive groves 
(28)Pastures 
(29)Peatbogs 
(30)Permanently irrigated land 
(31)Port areas 
(32)Rice fields 
(33)Road and rail networks and associated land 
(34)Salines 
(35)Salt marshes 
(36)Sclerophyllous vegetation 
(37)Sea and ocean 
(38)Sparsely vegetated areas 
(39)Sport and leisure facilities 
(40)Transitional woodland/shrub
(41)Vineyards 
(42)Water bodies 
(43)Water courses
Answer: 
\end{verbatim}

\textbf{Prompt for RGB-only input for BigEarthNet}
\begin{verbatim}
Instructions: Answer the question asked after the given image. Output format: Output the 
option numbers corresponding to the correct answer in the format "(X)" where X is the 
correct digit choice (among 1 to 43). In case of multiple correct answers, output the 
answer choices in the same format separated by commas. For example, if the correct 
answers are choices 1 and 3, output "(1),(3)". Only output the option numbers in the 
above format and DO NOT OUTPUT the full answer text or any other extra words.

Question: To which of the following class does the given image belong to? 
Select all that apply. Possible answer choices: 
(1)Agro-forestry areas 
(2)Airports 
(3)Annual crops associated with permanent crops 
(4)Bare rock 
(5)Beaches, dunes, sands 
(6)Broad-leaved forest 
(7)Burnt areas 
(8)Coastal lagoons 
(9)Complex cultivation patterns 
(10)Coniferous forest 
(11)Construction sites 
(12)Continuous urban fabric 
(13)Discontinuous urban fabric 
(14)Dump sites 
(15)Estuaries 
(16)Fruit trees and berry plantations 
(17)Green urban areas 
(18)Industrial or commercial units 
(19)Inland marshes 
(20)Intertidal flats 
(21)Land principally occupied by agriculture, with significant areas of natural vegetation 
(22)Mineral extraction sites 
(23)Mixed forest 
(24)Moors and heathland 
(25)Natural grassland 
(26)Non-irrigated arable land 
(27)Olive groves 
(28)Pastures 
(29)Peatbogs 
(30)Permanently irrigated land 
(31)Port areas 
(32)Rice fields 
(33)Road and rail networks and associated land 
(34)Salines 
(35)Salt marshes 
(36)Sclerophyllous vegetation 
(37)Sea and ocean 
(38)Sparsely vegetated areas 
(39)Sport and leisure facilities 
(40)Transitional woodland/shrub
(41)Vineyards 
(42)Water bodies 
(43)Water courses
Answer: 
\end{verbatim}

\textbf{Prompt for RGB-only input for BigEarthNet, 19 classes}
\begin{verbatim}
Instructions: Answer the question asked after the given image. Output format: Output the 
option numbers corresponding to the correct answer in the format "(X)" where X is the 
correct digit choice (among 1 to 19). In case of multiple correct answers, output the 
answer choices in the same format separated by commas. For example, if the correct 
answers are choices 1 and 3, output "(1),(3)". Only output the option numbers in the 
above format and DO NOT OUTPUT the full answer text or any other extra words.

Question: To which of the following class does the given image belong to? 
Select all that apply. Possible answer choices: 
(1)Agro-forestry areas
(2)Arable land
(3)Beaches, dunes, sands
(4)Broad-leaved forest
(5)Coastal wetlands
(6)Complex cultivation patterns
(7)Coniferous forest
(8)Industrial or commercial units
(9)Inland waters
(10)Inland wetlands
(11)Land principally occupied by agriculture, with significant areas of natural vegetation
(12)Marine waters
(13)Mixed forest
(14)Moors, heathland and sclerophyllous vegetation
(15)Natural grassland and sparsely vegetated areas
(16)Pastures
(17)Permanent crops
(18)Transitional woodland, shrub
(19)Urban fabric
Answer: 
\end{verbatim}

\textbf{Prompt for RGB-only input for EuroSat}
\begin{verbatim}
Instructions: Answer the question asked after the given image.
Output format: Output the option number corresponding to the correct answer in the 
format "(X)" where X is the correct number choice (among 1 to 10) for the single label 
classification task. Only output the option number in the above format and DO NOT
OUTPUT the full answer text or any other extra words.

Question: To which of the following class does the given image belong to?
Possible answer choices:
(1)AnnualCrop
(2)Forest
(3)HerbaceousVegetation
(4)Highway
(5)Industrial
(6)Pasture
(7)PermanentCrop
(8)Residential
(9)River
(10)SeaLake
Answer: 
\end{verbatim}

\section{Dataset details}

The experiments conducted in this study are done in evaluation-only Zero-Shot manner. We provide details about the datasets used.

%BigEarthNet -- 590,326
%Eurosat -- 27,000
%ForTy -- 78,644

% BigEarthNet has 2 resolutions -- RGB, False Color Composite Image, NDVI and NDWI have resolution 120x120 while the 2 NDMI images have resolution 60x60

The BigEarth dataset~\cite{BigEarthNet} has 590k images. The main inputs are of resolution 120x120 pixels, the last two images are 60x60 pixels. It is a multi-label dataset with more than one correct label per image. 
It has 43 classes. Because of ambiguity within the original version, there is an 19-class variant which is often used in the literature~\cite{BigEarthNet19Cl}.

The EuroSat dataset~\cite{helber2019eurosat} has 27k images with image resolution of 64x64 pixels. It has 10 classes.
\end{document}